\documentclass[11pt,a4paper]{article}
\usepackage[hyperref]{acl2020}
\usepackage{times}
\usepackage{latexsym}
\usepackage{xspace}

\usepackage{microtype}
\usepackage{amsmath}
\usepackage{multirow}
\usepackage{multicol}
\usepackage{paralist}

\usepackage{pifont}
\newcommand{\cmark}{\ding{51}}%
\newcommand{\xmark}{\ding{55}}%

\usepackage{enumitem}
\usepackage{subcaption}

\setitemize{noitemsep,topsep=0pt,parsep=0pt,partopsep=0pt}
\aclfinalcopy 

\newcommand{\method}{MC-Tailor\xspace}

\title{Do You Have the Right Scissors?\\ Tailoring Pre-trained Language Models via Monte-Carlo Methods}

\author{
	Ning Miao ~~ Yuxuan Song ~~ Hao Zhou ~~ Lei Li\\ 
	ByteDance AI lab \\ 
	\texttt{\{miaoning,songyuxuan,zhouhao.nlp,lileilab\}@bytedance.com} 
}

\date{}

\usepackage{graphicx}

\begin{document}
\maketitle

\begin{abstract}
It has been a common approach to pre-train a language model on a large corpus and fine-tune it on task-specific data. 
In practice, we observe that fine-tuning a pre-trained model on a small dataset may lead to  over- and/or under-estimation problem. 
In this paper, we propose \method, a novel method to alleviate the above issue in text generation tasks by truncating and transferring the probability mass from over-estimated regions to under-estimated ones. 
Experiments on a variety of text generation datasets show that \method consistently and significantly outperforms the fine-tuning approach. Our code is available at \url{https://github.com/NingMiao/MC-tailor}.
\end{abstract}

\section{Introduction}
\label{sec:introduction}
Recently, pre-trained language models~(PLM), \emph{e.g.} GPT-2~\citep{radford2019language}, have shown great promise in many applications of natural language generation, such as stylized text generation~\citep{syed2019adapting} and dialog system~\citep{DBLP:journals/corr/abs-1901-08149}. 
PLM is obtained by first \textit{pre-training} on large-scaled raw sentences~(always general domain corpus), and then used in downstream tasks by \textit{fine-tuning} on task-specific datasets~(always from some specific domains).
Specifically, given a pre-trained GPT-2 model, to generate sentences of email domain, we always need to fine-tune the GPT-2 on a small set of email domain corpus. 


However, we argue that to get desired sentence outputs, fine-tuning PLM on a specific domain dataset is not necessarily the best, especially when the fine-tuning dataset is of a small size. 
Typically, fine-tuning is conducted through Maximum Likelihood Estimation~(MLE), with which the resulting model distribution will be asymptotically consistent with true distribution when the fine-tuning dataset has infinite data samples. 
But it is not the case of fine-tuning on small datasets, which always leads to the mismatch problem of the real and model distributions.

\begin{figure}[t]
    \centering
    \includegraphics[width=1.0\columnwidth]{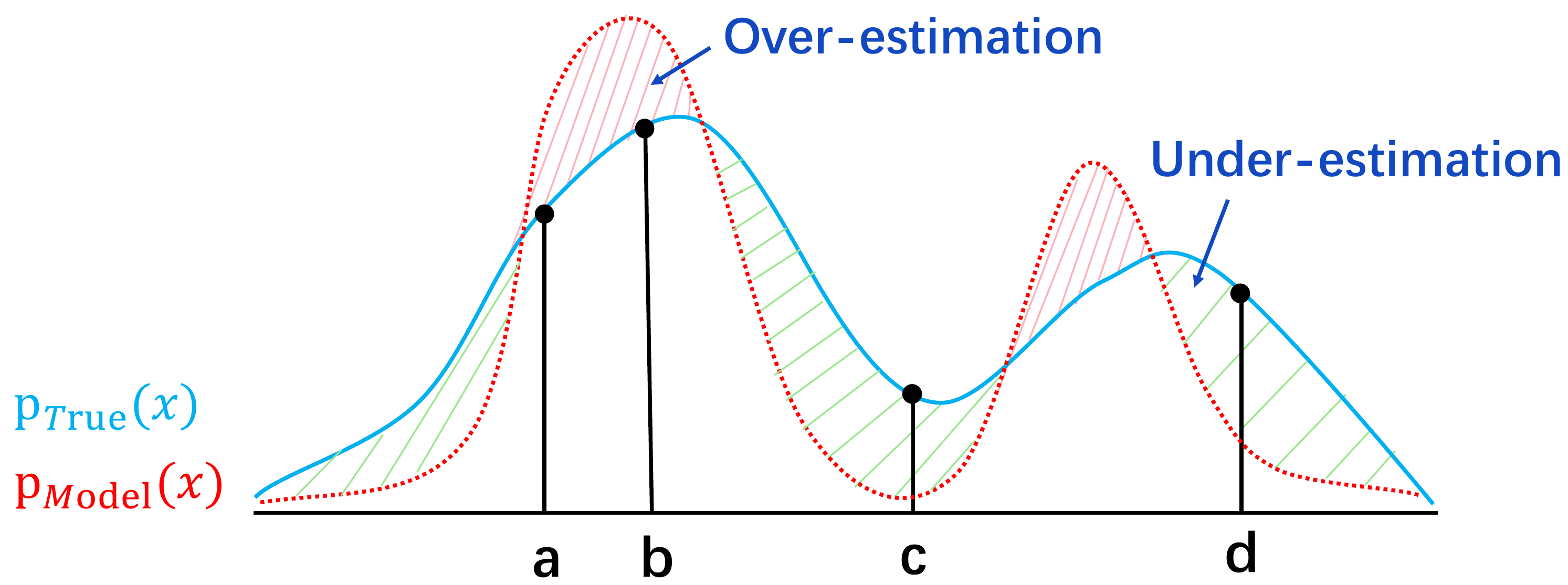}
    \caption{The over- and under-estimation problems of the model distribution. For example, sample \textbf{b} represents the simple sentence ``Yes .'', whose probability is over-estimated. Its model NLL~(4.01, negative log-likelihood) is significantly lower than the $95\%$ confidence interval of its real NLL [4.89, 5.37], which is estimated on the training set.}
    \label{fig:intro}
\end{figure}

Specifically, MLE minimizes the Kullback–Leibler~(KL) divergence between model and true distributions.
\citet{theis2015note} point out that minimizing KL avoids assigning an extremely small probability to any data point but assigns a lot of probability mass to non-data regions, which leads to a gap between $P_{Real}$ and $P_{Model}$.
Additionally, simple data patterns in the fine-tuning dataset could be easily memorized and over-estimated. Meanwhile, the complex ones may be under-estimated. 
The above problem is not severe with adequate data samples, but non-trivial when the size of the fine-tuning dataset is not large enough.~({see Figure \ref{fig:intro}}).

To address the over- and under-estimated problem, in this paper, we propose \method, which can tailor the resulting density of model distribution by cutting the probability mass of over-estimated zones to under-estimated zones, leading to more realistic model distribution after fine-tuning.
Concretely, \method consists of two components: a ratio estimator to distinguish over- and under-estimated regions of model distribution; and an early rejection sampling~(ERS) component to tailor~(reassign) probability mass and efficiently obtain sampled sentences from the model distribution.
Note that the proposed ERS is inspired by Sequential Monte Carlo~(SMC, \citet{doucet2000sequential}), but can avoid the degeneration from SMC, as it directly kills samples rather than performs resampling.

We conduct experiments on various data sets to verify the effectiveness of the proposed \method. 
Empirical results show that \method can generate significantly better samples than finetuning, and the resulting model distributions of our model are closer to real data distributions.

\section{Pre-Trained Language Model}


Language models generally estimate the density of sentences in real context within an autoregressive style:
\begin{equation}
    P(x)=\prod_{i=1}^{N} P(x_i|x_{[1:i-1]}),
\end{equation}
where $x$ is a sentence with length $N$.
Recently, with an extremely large number of parameters, pre-trained language models like GPT-2~\citep{radford2019language} and Transformer-XL~\citep{dai2019transformer} have shown great promise in text generation. PLMs are first trained on a huge general domain data set and then fine-tuned on specific domain datasets of different downstream tasks.

Specifically, given a pre-trained GPT2 model, to generate sentences of email domain, we always need to fine-tune the GPT2 on a small set of email domain corpus.
Additionally, PLMs have some other important applications.
\citet{miao2019cgmh} use fine-tuned language models for constrained text generation. \citet{DBLP:journals/corr/abs-1901-08149} fine-tune GPT-2 on a dialog data set to boost the performance of dialog system.

However, as stated in the Introduction, directly fine-tuning the PLM on a small dataset may lead to the mismatch problem, namely the over- and under-estimated problem between the true distribution and the model distribution.
In the next section, we propose a new method to alleviate this problem.

\section{Proposed \method}
To mitigate the above shortcomings of finetuning, we propose \method , which generates samples from a modified sample distribution. \method is composed of a ratio estimator, which detects over- and under-estimate regions of model distributions, and the Early Rejection Sampling algorithm~(ERS), which accelerates sampling while ensuring sample quality.
\subsection{Ratio Estimator}
Ratio estimator is a common technique to measure the gap between two related distributions~\citep{improving}. In this work, We apply \textbf{ratio estimator~$\gamma(x)$} to estimating $\frac{P_{\text{Model}}(x)}{P_{\text{True}}(x)}$, the probability ratio of sentence $x$ in fine-tuned model distribution $P_{\text{Model}}(x)$ and true distribution $P_{\text{True}}(x)$. To tailor the probability from a finetuned PLM, we cut the probabilities of over-fitting samples. Specifically, when $\gamma(x)>1$, i.e., the model over-estimates the probability of sample $x$, we remove $x$ with a probability of $1-\frac{1}{r(x)}$ to approximate $P_{\text{True}}(x)$. After normalization, probabilities of under-estimated areas will increase correspondingly.
The resulting new distribution is $P_{\text{Tailor}}\propto \frac{P_{\text{Model}}(x)}{max(\gamma(x), 1)}$.  
In this work, we try several different structures of ratio estimators.

\noindent\textbf{Convolutional Ratio Estimator.}
Since ratio estimation shares similar properties with classification problems and convolutional neural networks~(CNN) are powerful classifiers, our first thought is to build a CNN-based ratio estimator.
To be concrete, we use a two-layer CNN to predict whether $x$ is from true or learned distribution. By training with cross-entropy loss,
\begin{equation}
    Softmax(\text{CNN}(x))\xrightarrow{} \frac{P_{\text{Model}}(x)}{P_{\text{True}}(x)+P_{\text{Model}}(x)}.
\end{equation}
Naturally, we define
\begin{equation}
    \gamma(x)=\frac{Softmax(\text{CNN}(x))}{1-Softmax(\text{CNN}(x))}.
\end{equation}
\begin{figure*}[t!]
    \centering
    \begin{subfigure}{0.25\textwidth}
        \centering
        \includegraphics[width=\textwidth]{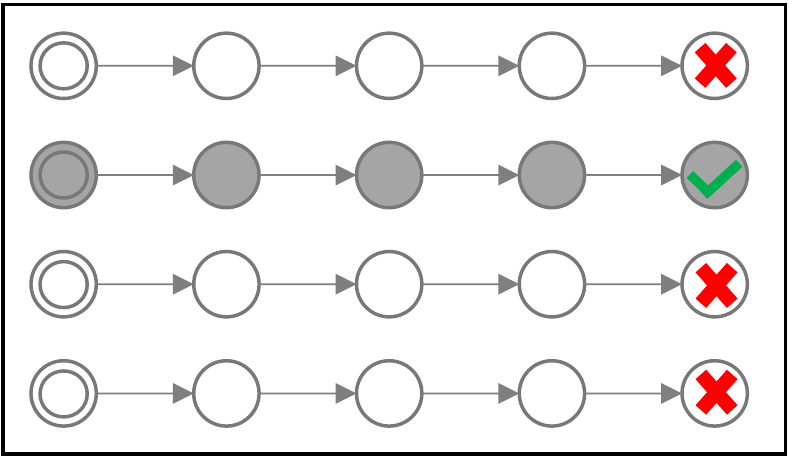}
        \caption{RS}
        \label{fig:model_RS}
    \end{subfigure}%
    ~ 
    \begin{subfigure}{0.25\textwidth}
        \centering
        \includegraphics[width=\textwidth]{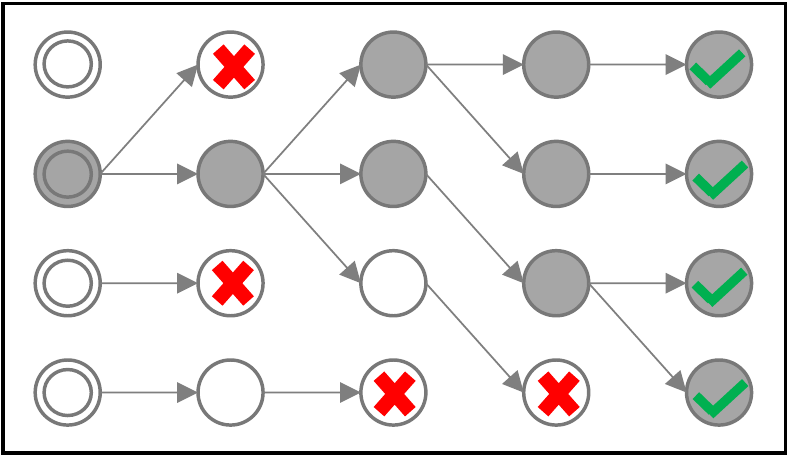}
        \caption{SMC}
    \end{subfigure}
    ~
    \begin{subfigure}{0.25\textwidth}
        \centering
        \includegraphics[width=\textwidth]{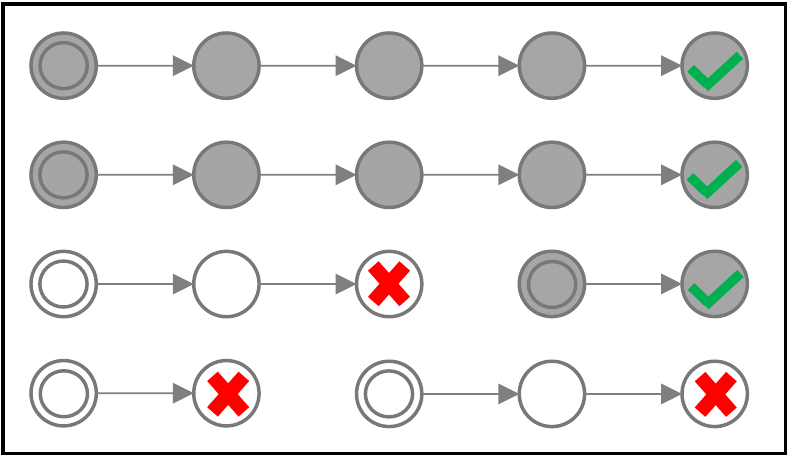}
        \caption{ERS}
    \end{subfigure}
    \label{model}
    \caption{Illustration of three sampling algorithms. Concentric circles are newly born particles. Green checkmarks and Red crosses appear when particles are accepted and killed, respectively. Gray circulars represent particles finally accepted while white circulars stand for the opposite.}
\end{figure*}


\noindent\textbf{Dual Ratio Estimator.}
Though the basic convolutional ratio estimator is easy to apply, it makes sampling inefficient. For most sentence $x$, we can roughly predict whether it is in a specific domain or suffering from over-estimation by the first a few words.
However, $\gamma(x)$ can only be obtained after a full sentence is generated, so massive computing resources are wasted on generating unpromising samples.

To determine whether a prefix $x_{[1:i]}$ is promising, we can estimate 
\begin{equation}
    \gamma^{'}(\hat{x}_{[1:i]})=\min_{x_{[1:i]}=\hat{x}_{[1:i]}}(\gamma(x)),
\end{equation}
where $\gamma^{'}(\hat{x}_{[1:i]})$ is the minimum ratio of all sentences with prefix $\hat{x}_{[1:i]}$. If $\gamma^{'}(\hat{x}_{[1:i]})$ is greater than a pre-defined threshold, all sentences with prefix $x_{[1:i]}$ should be rejected. As a result, we do not need to waste time to continue sampling.

But if we directly train $\gamma^{'}(\hat{x}_{[1:i]})$ to distinguish $P_{True}(x_{[1:i]})$ from $P_{Model}(x_{[1:i]})$, we will end up getting the average value of $\gamma (x)$ for all sentences with prefix $x_{[1:i]}$, rather than the minimum value. If so, some sentences with low $\gamma (x)$ will be erroneously rejected. Luckily, the properties of min-max dual sheds some light on this problem. We first define $\gamma^{''}(x)=\max_i( \gamma^{'}(x_{[1:i]}))$ as the dual form of $\gamma^{'} (x)$. Under some weak conditions, we can prove that if $\gamma^{''}(x)$ approximates $\frac{P_{\text{Model}}(x)}{P_{\text{True}}(x)}$, then $\gamma^{'}(\hat{x}_{[1:i]})$ approximates $\min(\gamma(x))$ for $x$ with prefix $x_{[1:i]}$. Similar to training $\gamma (x)$, we train $\gamma^{''}(x)$ by distinguishing  $P_{\text{True}}(x)$ from $P_{\text{Model}}(x)$. Since $\gamma^{''}(x)$ is a function of $\gamma^{'}(\hat{x}_{[1:i]})$, we can get a set of proper parameters for $\gamma^{'}(\hat{x}_{[1:i]})$.


\noindent\textbf{Hierarchical Ratio Estimator.}
Since a single ratio estimator may not be powerful enough to accurately estimate $\frac{P_{\text{Model}}(x)}{P_{\text{Real}}(x)}$, we break down the workload to several $\gamma_i(x)$ in the spirit of boosting. We first train $\gamma_0(x)$ to estimate $\frac{P_{\text{Model}}(x)}{P_{\text{Real}}(x)}$, and get $P_{\text{Tailor}}^0(x)$. And then we use $\gamma_1(x)$ to estimate the gap between $P_{\text{Real}}$ and $P_{\text{Tailor}}^0(x)$... With the collaboration of $\gamma_i(x)$, we can get a more accurate $P_{\text{Tailor}}^n(x)$. 
Using hierarchical ratio estimators also avoids using a single but complicated ratio estimator, which is prone to over-fitting. Similarly, we can add hierarchy to the dual ratio estimator to make a hierarchical dual ratio estimator.



\subsection{Efficient Sampling}
\label{sec:sampling}
In this part, we introduce our specially designed Early Rejection Sampling~(ERS) algorithm for \method. Improved from Sequential Monte Carlo, ERS can efficiently generate samples with high diversity.

\noindent\textbf{Rejection Sampling}
By applying RS, we first generate a batch of samples from $P_{\text{Model}}$, and then rejecting some samples by rejection ratio $1-\frac{1}{max(\gamma(x),1)}$.
However, RS is very inefficient in actual use since it rejects samples at the end of sampling. As shown in Figure~\ref{fig:model_RS}, lots of computation resources are wasted on ultimately rejected samples.

\noindent\textbf{Sequntial Monte Carlo}
Instead of rejecting samples at the end of sampling, SMC performs resampling at each step. The unnormalized resampling weight at step $i$ is provided by $ \frac{\gamma^{'}{(x_{[1:i-1]})}}{\gamma^{'}{(x_{[1:i]})}}$, leading to an asymptotically unbiased estimator.
However, SMC suffers from serious degeneracy problem. In other words, samples from SMC tend to share a very small number of the ancestors because most of the ancestors are killed during resampling. As a result, sample diversity of SMC is critically low. 

\noindent\textbf{Early Rejection Sampling}
To overcome the degeneracy problem of SMC and increase sample diversity. We propose Early Rejection Sampling~(ERS) algorithm. ERS first uniformly samples a real number $r$ in $(0,1)$. After step $i$, if $\gamma^{'}(x[1:i])>\frac{1}{r}$, this particle is killed immediately and computation resources are released to parallel threads.
The main difference between ERS and RS is that ERS kills unpromising particles before they are fully generated. But unlike SMC, there is no correlation between ERS samples, resulting in higher sample diversity.

\section{Experiments}
\label{sec:experiment}
In this section, We empirically compare the sample quality of our model and baseline models. We first set up experiments and show results in Section~\ref{sec:experiment_results}. 
\begin{table*}[htb]
    \centering
    \footnotesize
    \begin{tabular}{ll|c|c|c|c|c} 
        \hline
        Datasets& &\#Train& Style&  Fine-tune&$\text{\method}_{\text{RS}}$&$\text{\method}_{\text{ERS}}$ \\\hline
        Ontonotes& &&&&\\
        &-bn &12k &Broadcast news& 124&117&\textbf{111} \\
        &-bc & 12k &Broadcast dialog & 268&\textbf{144}&153 \\
        &-mz & 7k &Magazine & 126&112&\textbf{110} \\
        &-nw & 35k &Newswire & 111&110&\textbf{100} \\
        &-tc & 13k &Telephone dialog& 140&136&\textbf{134} \\
        &-wb & 17k &Web & 166&138&\textbf{136} \\\hline
        Switchboard &&203k&Formal dialog&198&\textbf{165}&169\\\hline
        DailyDialog &&76k&Daily dialog&120&117&\textbf{113}\\\hline
        IWSLT-16 &&133k&Conference speech&240&217&\textbf{213}\\\hline

    \end{tabular}
        \caption{Rev-PPL of each method. All methods start from the same pre-trained GPT2 model. $\text{\method}_{\text{RS}}$ represents single-layer \method with rejection sampling and $\text{\method}_{\text{ERS}}$ is a hierarchical \method with 3 layers and ERS algorithm. Results of SMC are not reported since it leads to very poor Rev-PPLs because of the lack of sample diversity.}
    \label{tab:rev-ppl}
\end{table*}

\subsection{Experimental Setup}
We conduct experiments on 9 data sets with different styles and sizes. And we use five different metrics, including human evaluation, to measure the generation performance of each method.

\noindent\textbf{Datasets.} We use the following data sets for experiments.
\begin{itemize}
    \item \textbf{Ontonotes}~\citep{pradhan2013towards} is a multi-genre data set for sequence annotation. We use sentences from six genres~(bn, bc, mz, nw, tc, wb) for the experiment.
    \item \textbf{Switchboard}~\citep{Jurafsky-etal:1997} and \textbf{DailyDialog}~\citep{li2017dailydialog} are large and medium scale dialog data sets, of which only responses are used for the experiment.
    \item \textbf{IWSLT-16}~\citep{cettolo2016iwslt} is a data set of paired conference speeches for machine translation. We use English sentences from De-En pairs to test model performance on the special conference speech domain.
\end{itemize}

\begin{table*}[h]
    \centering
    \footnotesize
    \begin{tabular}{c|l|c|c} 
        \hline
        Refs&Sentences &NLL~(Fine-tune)&NLL~($\text{\method}_{\text{ERS}}$)\\\hline
        \textbf{a}&Thank you everyone for watching .&18.03&18.65\\\hline
        \textbf{b}&Yes . &4.01&4.77\\\hline
        \textbf{c}&What does that mean in the context of your book ?&26.56&26.44\\\hline
        \textbf{d}&And it did n't hurt too badly .&23.24&22.97\\\hline
    \end{tabular}
        \caption{NLL comparison of $\text{\method}_{\text{ERS}}$ and the baseline on Ontonotes-bc. $\text{\method}_{\text{ERS}}$ successfully reallocates the probabilities of over-estimated samples~(simple sentences such as \textbf{a} and \textbf{b}) to under-estimated ones~(complicated sentences such as \textbf{c} and \textbf{d}). }
    \label{tab:NLL}
\end{table*}

\begin{table*}[h]
    \centering
    \footnotesize
    \begin{tabular}{l|l|l} 
        \hline
        Methods&Fine-tune&$\text{\method}_{\text{ERS}}$\\\hline
        \multirow{5}{*}{Samples}&Right .&She should be charged with rape .\\\cline{2-3}
        &In the case if you think of this -&And do you still feel that way every day ?\\\cline{2-3}
        &Oh well .&But it would be tough .\\\cline{2-3}
        &I 've been there n't said anything wrong .&He knew about the attack at the Paris offices .\\\hline
    \end{tabular}
        \caption{Generated samples of each method on Ontonotes-bc. Samples from $\text{\method}_{\text{ERS}}$ are more informative and coherent with the target style than the baseline method.}
    \label{tab:example}
\end{table*}

\noindent\textbf{Evaluation Metrics.} To evaluate the generation quality and diversity, we use the following metrics.
\begin{itemize}
    \item \textbf{PPL} reflects the average density of samples from test set in a generative model. Models with lower PPLs have more similar model distributions with real contexts. Unlike baseline models, \method only has an unnormalized log-probability. We estimate the normalization constant of \method by importance sampling and calculate PPLs directly from the normalized log-probability.
    \item \textbf{Rev-PPL} is a good indicator for both sample quality and diversity, which is derived by first training a language model with generated samples and calculating the PPL of test set in the language model.
    \item \textbf{EMD-l} is the earth mover distance between sentence lengths of real and generated data.
    \item \textbf{EMD-f} is the earth mover distance between word frequencies of real and generated data.
    \item \textbf{Human Evaluation Score} is added to reflect the comprehensive sample quality. We ask 4 volunteers to select a score from \{0, 0.5, 1\} for each sample according to their fluency and coherence with the target style. In 85\% cases, at least three volunteers give the same score, showing the reliability of the human evaluation.
\end{itemize}

\noindent\textbf{Model Details.}
In all the experiments, we use the released {GPT-2} with 117M parameters as the pre-trained language model. We first fine-tune {GPT-2} on each dataset and then build our tailor on it. Early-stop is applied to avoid over-fitting.  For ratio estimators, we use simple CNNs with two convolution layers where (filter\_number, kernel\_size) is set to (10,5) and (5,5), respectively.


\subsection{Experimental Results}
\label{sec:experiment_results}

    
Rev-PPLs of different models are shown in Table~\ref{tab:rev-ppl}. 
We find that \method significantly reduces Rev-PPLs than fine-tuning baseline in data sets of different sizes, from Ontonotes-mz with only 7k training samples to relatively large Switchboard data set with more than 200k samples.
We also notice that multi-layer $\text{\method}_{\text{ERS}}$ performs better than single-layer $\text{\method}_{\text{RS}}$, which confirms the point in Section~\ref{sec:sampling} that the gap between $P_{\text{Model}}$ and $P_{\text{Data}}$ is too complex for a single-layer ratio estimator to estimate. Sample NLLs of each method (Table~\ref{tab:NLL}) further confirms that \method succeeds in decreasing the probabilities of over-estimated simple patterns and reallocating them to under-estimated samples.

We further compare \method with the baseline model under other metrics. From table~\ref{tab:scores}, we find \method greatly reduce
PPL, which means increased probabilities of generating samples similar to test samples. And we can draw the conclusion that sample distributions of \method are closer to real sample distributions, with lower EMD-l and EMD-f. What's more, human evaluation scores of \method are about 10\% higher than fine-tuning, which indicates better sample quality to human eyes. Cases shown in Table~\ref{tab:example} further demonstrate the advantage of \method in fluency and informativeness.  
Seq-GAN is also compared in our experiment. However, rev-ppls of GANs are even higher than directly fine-tuning GPT-2, and they are especially difficult to train. So we remove Seq-GAN from baseline models.

The acceleration effect of ERS is also verified in the experiment. For \method with 1, 2, and 3 layers of ratio estimator, ERS reduces 30\%, 79\%, and 90\% of computation wasted on unpromising samples, achieving 1.5x, 2.8x, 5x accelerations, respectively.

%



\begin{table}[t]
    \centering
    \footnotesize
    \begin{tabular}{l|c|c|c|c|c}
        \hline
        Data & MCT  & PPL &  EMD-l & EMD-f &Human\\\hline
        \multirow{2}{*}{Onto-bn} & \xmark  &  34.1 & 4.31 & 0.57 & 0.60\\\cline{2-6}
         & \cmark  & \textbf{30.1}  & \textbf{1.90} & \textbf{0.53}&\textbf{0.81} \\\hline
         \multirow{2}{*}{Onto-bc} & \xmark  & 30.9  & 6.74 & 0.67 & 0.40\\\cline{2-6}
         & \cmark  & \textbf{23.1}  & \textbf{1.62} & \textbf{0.55}&\textbf{0.67} \\\hline
         \multirow{2}{*}{Onto-mz} & \xmark  & 43.4  & 5.60 & 0.69 & 0.71\\\cline{2-6}
         & \cmark  &  \textbf{39.7} & \textbf{3.33} & \textbf{0.64}& \textbf{0.76}\\\hline
         \multirow{2}{*}{Onto-nw} & \xmark  &  37.0 & 4.94 & 0.61 & 0.65 \\\cline{2-6}
         & \cmark  & \textbf{36.1}  & \textbf{3.66} & \textbf{0.54}&\textbf{0.70} \\\hline
         \multirow{2}{*}{Onto-tc} & \xmark  &  24.8 & 4.19 & \textbf{0.64} & 0.54 \\\cline{2-6}
         & \cmark  & \textbf{23.8}  & \textbf{2.46} & \textbf{0.64}& \textbf{0.54} \\\hline
         \multirow{2}{*}{Onto-wb} & \xmark  & 60.9  & 3.31 & 0.61 & 0.46 \\\cline{2-6}
         & \cmark  &  \textbf{52.8} & \textbf{2.40} & \textbf{0.51}& \textbf{0.60}\\\hline
         \multirow{2}{*}{SB} & \xmark  &  19.7 & 8.75 & 0.60 & 0.48\\\cline{2-6}
         & \cmark  & \textbf{18.9}  & \textbf{5.21} & \textbf{0.51}& \textbf{0.54}\\\hline
         \multirow{2}{*}{DD} & \xmark  & 30.3  & 5.25 & 0.47 & 0.60\\\cline{2-6}
         & \cmark  & \textbf{29.1}  & \textbf{3.32} & \textbf{0.45}& \textbf{0.62}\\\hline
         \multirow{2}{*}{IWSLT} & \xmark  & 23.3  & 5.21 & 0.61 & 0.32\\\cline{2-6}
         & \cmark  &  \textbf{20.9} & \textbf{2.99} & \textbf{0.55}& \textbf{0.40}\\\hline

    \end{tabular}
    \caption{PPL, EMD-l, EMD-f and human evaluation score of $\text{\method}_{\text{ERS}}$ with 3 layers and fine-tuning. MCT means whether to use our proposed \method or to direct fine-tune. SB and DD represent the Switchboard and DailyDialog data sets, respectively. By one-tail t-tests, we find that improvements in human evaluation scores are significant, with p-values smaller than 0.05.}
    \label{tab:scores}
\end{table}

\section{Conclusion}
In this paper, we propose \method to alleviate the over- and under-estimation problem between true and model distributions.
\method is composed of a ratio estimator, which adjusts the probabilities of MLE fine-tuned PLMs to approximate true distributions, and the ERS to accelerate sampling while ensuring sample quality. Experiments on various datasets show the effectiveness and efficiency of \method .

\bibliography{acl2020}
\bibliographystyle{acl_natbib}
\end{document}